\documentclass{article} 
\usepackage{iclr2027_conference,times}

\usepackage{amsmath,amsfonts,bm}

\def\eqref#1{equation~\ref{#1}}

\def\1{\bm{1}}

\DeclareMathAlphabet{\mathsfit}{\encodingdefault}{\sfdefault}{m}{sl}
\SetMathAlphabet{\mathsfit}{bold}{\encodingdefault}{\sfdefault}{bx}{n}

\usepackage{hyperref}
\hypersetup{hidelinks}
\usepackage{url}
\usepackage{booktabs}
\usepackage{array}
\usepackage{graphicx}
\usepackage{wrapfig}
\usepackage{adjustbox}
\usepackage{changepage}
\usepackage{enumitem}
\usepackage{xspace}
\usepackage{tcolorbox}
\tcbuselibrary{listings,breakable,skins}
\usepackage{placeins}

\title{Beyond Skill Evolution: \\Self-Evolving Context Management Policies for Long-Horizon Agent Harnesses}

\author{
  \textbf{Weiyuan Li\textsuperscript{1,2$\ast$}},
  \textbf{Jinghan Xu\textsuperscript{1,2$\ast$}},
  \textbf{Aili Chen\textsuperscript{2,3}},
  \textbf{Xintao Wang\textsuperscript{2,3}}, \\
  \textbf{Shuang Liang\textsuperscript{1,2}},
  \textbf{Jiaqing Liang\textsuperscript{1,2}},
  \textbf{Deqing Yang\textsuperscript{1,2$\dagger$}},
  \\
  \textsuperscript{1}School of Data Science, Fudan University,\\
  \textsuperscript{2}Shanghai Key Laboratory of Data Science,\\
  \textsuperscript{3}College of Computer Science and Artificial Intelligence, Fudan University\\
  \textsuperscript{$\ast$}Equal contribution; \textsuperscript{$\dagger$}Corresponding author\\
  \texttt{weiyuanli25@m.fudan.edu.cn},
  \texttt{jhxu25@m.fudan.edu.cn},
  \texttt{yangdeqing@fudan.edu.cn}
}

\newcommand{\methodname}{ContextEvo}
\newcommand{\method}{\textsc{\methodname}} 

\arxivcopy
\begin{document}

\maketitle

\begin{abstract}
    Harness evolution improves LLM agents by learning from execution trajectories, but existing experience- and skill-based methods are less effective on long-horizon tasks. 
    As interactions grow, useful evidence can be buried by redundant or outdated context, making context management itself a key bottleneck. 
    We introduce \method{}, a framework that learns a context policy from long-horizon trajectories. 
    \method{} reconstructs the model-visible context at key decision points, identifies context-related failures, and applies targeted policy updates. 
    Starting from the open-source Pi-agent harness, \method{} improves performance across three long-horizon task benchmarks, achieving results comparable to or better than several prominent agent harnesses, including Codex, OpenCode, and OpenClaw.
    Additional analyses show that fixed or locally evolved context strategies can fall short under long-horizon information pressure, while our methods adapt to the information demands of each environment.
    
\end{abstract}

\begin{figure}[!b]
    \centering
    \includegraphics[width=\linewidth]{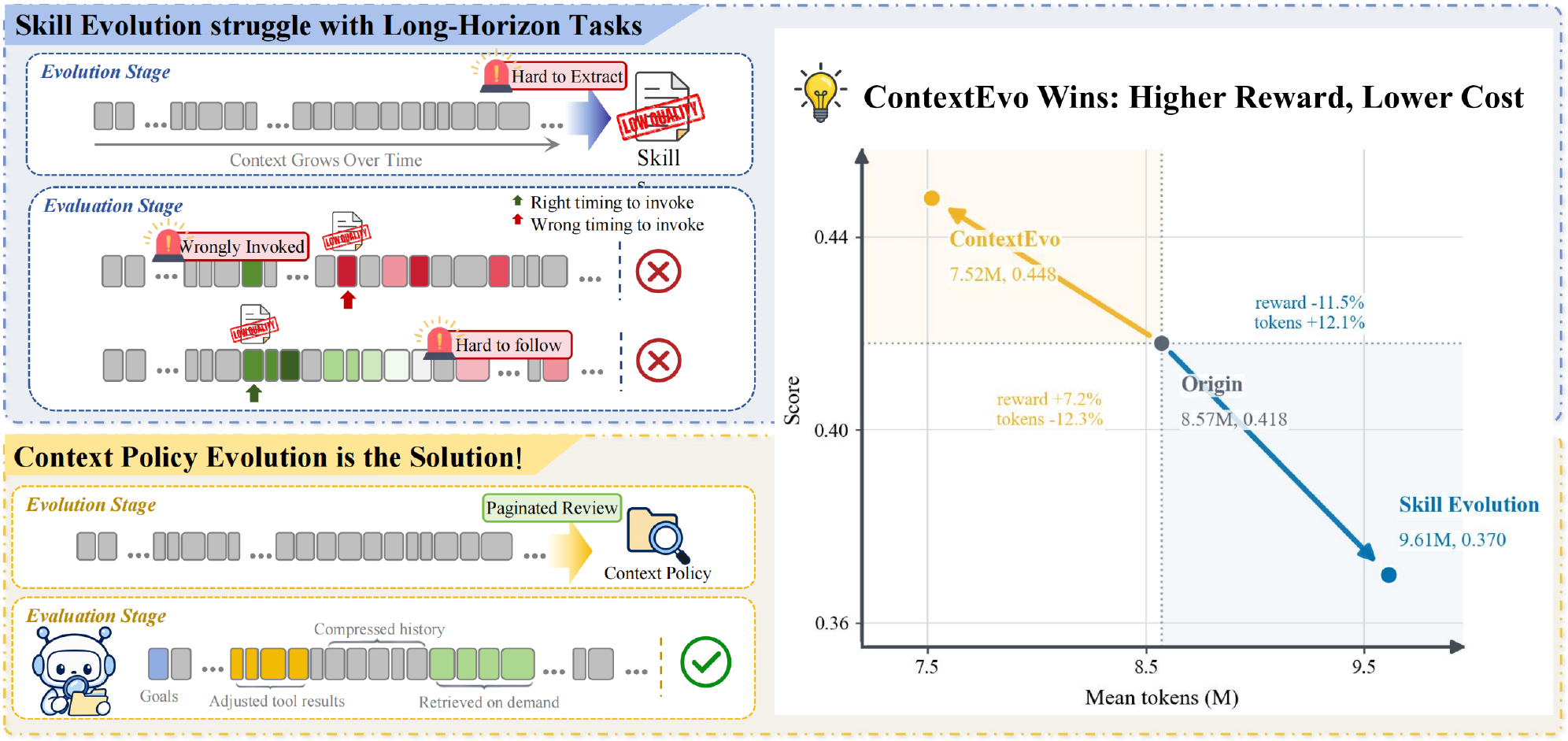}
    \caption{\textbf{From skill evolution to context-policy evolution.}
As interaction histories grow, \textbf{\emph{skill evolution}} may fail at both stages: extracting low-quality skills during evolution and invoking them at the wrong time or following them unreliably during evaluation. 
In contrast, our \textbf{\method{}} evolves an efficient context policy, achieving higher scores with fewer tokens.}
    \label{fig:main}
\end{figure}

\section{Introduction}
LLM agents are increasingly capable of solving complex, long-horizon tasks, including web navigation, software engineering, and computer use \citep{zhou2024webarena,jimenez2024swebench,yuan2026osworld2}.
In these tasks, strong agent performance depends not only on a capable base model, but also on a well-designed harness that manages context, tool use, and execution \citep{lin2026ahe,lee2026metaharness}.
As static harnesses may struggle to adapt to different task requirements, recent work has explored harness evolution, where agent trajectories are used as feedback signals to improve harness components \citep{lee2026metaharness,zhang2026selfharness,lin2026ahe}.

A main direction of harness evolution is to identify reusable experience from past interactions and incorporate it into agent harnesses, often in the form of rules, prompts, or skills \citep{alzubi2026evoskill,zhang2026evoskills,yang2026skillopt,agrawal2026gepa}.
While these approaches yield improvements in many scenarios, they are less effective in long-horizon tasks (Figure~\ref{fig:main}), where overlong context remains a major bottleneck \citep{liu2024lost,bai2024longbenchv2,min2026trace}.
In this setting, long interactions gradually accumulate redundant information, hindering effective skill invocation and degrading skill quality.

Previous studies have also explored approaches for context compression, organization, and memory management \citep{hu2025hiagent,kang2025acon,zhou2026mem1}.
Yet context management typically remains fixed or is only slightly adjusted even as the harness evolves, limiting adaptation across task environments \citep{yao2026arc,li2026acm, zhang2026selfharness, lee2026metaharness}.
This raises a critical question: \textbf{can long-horizon agents evolve their context policies automatically, systematically, and effectively?}

Evolving context policies, however, introduces unique challenges beyond existing skill or rule evolution approaches.
Context policies govern how information is presented, organized, and shared throughout task execution, making their effects hard to identify and, consequently, hard to optimize.
The first challenge is \textbf{failure attribution}: a failed trajectory does not directly reveal whether the problem comes from the model's reasoning, the environment, or the way information is managed, as these factors are often intertwined \citep{zhang2026selfharness,li2026growingharness}.
The second challenge is \textbf{target identification}: even when context management is identified as a source of failure, it remains unclear which part of the context policy should be evolved
\citep{lee2026metaharness,ma2026longhorizonharness,li2026growingharness}.

We introduce \method{}, a framework for evolving context policies across the full information lifecycle of long-horizon agent execution. 
From execution trajectories, \method{} reconstructs how information enters, persists, and reaches the model. 
It then uses cross-case evidence and counterfactual replay to determine whether a failure results from context management. 
When supported, this attribution is translated into a constrained policy update. 
By jointly evolving \textit{Input Assembly}, \textit{History Maintenance}, and \textit{Context Orchestration}, \method{} extends harness evolution from reusable rules, prompts, and skills to systematic management of information throughout the task.


We evaluate \method{} on three long-horizon benchmarks. 
Starting from Pi-agent, \method{} improves full-benchmark scores by 3.1--4.4 percentage points under DeepSeek-V4-Flash, and the evolved policies achieve competitive performance compared to other agent harnesses. 
Further analysis shows that the policy updates adapt to the information demands of each environment. Compared with existing methods that rely on fixed or locally adapted context-management strategies, \method{} achieves stronger performance by evolving the context policy across the full information lifecycle.

Our key contributions are as follows:
\begin{itemize}[leftmargin=*]
    \item We identify full-lifecycle context management as a critical target for long-horizon harness evolution, where existing evolution methods can struggle as context grows.
    \item We formulate context management as an evolvable full-lifecycle policy and develop \method{}, which uses evidence-based diagnosis and counterfactual validation to localize context failures and perform constrained policy updates.
    \item Across three long-horizon benchmarks, \method{} improves full-benchmark performance and produces policy updates that adapt to environment-specific information demands.
\end{itemize}

\section{Related Work}

\paragraph{Agent Performance on Long-Horizon Tasks.}
In long-horizon tasks, agents continuously interact with the environment and accumulate information throughout execution \citep{liu2024lost,bai2024longbenchv2,min2026trace}.
Previous work has improved agents through reinforcement learning \citep{chen2025rlinteractive,xi2025agentgymrl} and through the design of the agent harness, which organizes execution and controls what information the model receives across steps \citep{lee2026metaharness,ma2026longhorizonharness}.
At the harness level, methods use plans and environmental feedback to guide execution \citep{erdogan2025planandact,zhou2024lats}, or manage interaction histories to preserve relevant information \citep{hu2025hiagent,wan2026compass,kang2025acon}.
Yet a harness designed in advance may not meet all the needs that arise during a long-horizon task.
We therefore study how context management within the harness can evolve from execution trajectories.

\paragraph{Context Management for Complex Tasks.}
In complex real-world tasks, decisions often depend on goals, constraints, and observations encountered many steps earlier, while new interactions continually add to the history.
Context management is therefore critical to keeping relevant information available when it is needed.
Some methods organize or summarize the history around subgoals and completed steps \citep{hu2025hiagent,wu2026resum,ye2026agentfold}.
Others train agents to maintain compact task states \citep{zhou2026mem1,li2026mempo} or to choose memory operations during execution \citep{zhang2026memact,yu2026agenticmemory,lu2026contextcompression}.
External managers and persistent stores provide another way to retain information and present it when needed \citep{yi2026adacom,li2026acm,wu2026proactivememory,lin2026scroll}.
ACON \citep{kang2025acon} and TRACE \citep{min2026trace} further use execution feedback to refine context-compression rules.
Rather than optimizing a single memory operation or compression policy, \method{} uses execution trajectories to reveal deficiencies in the context-management mechanism and revise the harness: what enters the context, how information is retained or transformed, and which models receive it.

\paragraph{Self-Improving Agents and Harness Evolution.}
Self-improving agents use execution experience to update reusable playbooks, skills, memory, and other harness components \citep{zhang2026ace,zhang2026memskill,zhang2026selfharness,lin2026ahe,wei2026evoharness,jiang2026harnessevolve,xu2026harnesslens,li2026growingharness}.
Whether these updates help on later tasks also depends on how past experience is retrieved and used \citep{hu2026experiencereuse,ferraz2026retrievalagents}.
In long-horizon tasks, growing histories and irrelevant content can bury important information within the context \citep{wu2026resum,kang2025acon}.
Even a useful skill may therefore have little effect if the agent cannot use it when needed.
Failure-analysis methods examine trajectories to locate critical errors \citep{barke2026agentrx,zhu2025agentdebug,qi2026trajdebug}, while Causal Agent Replay \citep{shah2026car} tests the effects of changes to individual steps.
Despite these advances, it remains unclear how to systematically evolve context policies from trajectory-level evidence across the full information lifecycle.


\section{Preliminaries}
\label{sec:preliminaries}

This section introduces the setting of agent harness evolution and defines the \emph{context policy} studied in this work.
We first describe a general harness evolution setting, where the harness evolves while keeping the base model fixed.
We then focus on harness context management and define context policy to specify how information is presented, organized, and shared during agent execution.

\subsection{Agent Harness Evolution}

We consider an agent as a system composed of a base model \(M\) and a harness \(H\).
The base model is responsible for decision making, while the harness defines the execution procedure for the model, including interaction with external tools, environments, and other agents. More importantly, the harness determines the context upon which the model's decisions rely.

In this work, we focus on harness-level evolution: the base model \(M\) remains fixed, while the harness \(H\) evolves to improve agent performance.


\subsection{Context Policy}

Since the harness determines the context provided to the model during execution, we study harness evolution from the perspective of context management.
Within the harness, we define a \textbf{\emph{context policy}} \(P\) that governs how information is presented, maintained, and shared through the harness. 
At time \(t\), let \(\mathcal{I}_{\leq t}\) denote the information that has entered the system up to that point, including task instructions, observations, tool outputs, and intermediate information. 
The policy produces the model-visible context \(c_{t}=P(\mathcal{I}_{\leq t})\), with
\begin{equation}
    a_t \sim M(\cdot \mid c_{t}).
\end{equation}

A context policy is characterized along three dimensions: \emph{Input Assembly}, \emph{History Maintenance}, and \emph{Context Orchestration}.
Table~\ref{tab:context_policy} introduces these dimensions and common realizations in agent harnesses.
The policy space is not limited to these realizations and allows other implementations and combinations across dimensions.
Together, these dimensions define an abstraction of the full information lifecycle without depending on particular harness components.

\begin{table}[h]
\caption{Context-policy dimensions and common realizations in agent harnesses.}
\label{tab:context_policy}
\centering
\small
\setlength{\tabcolsep}{6pt}
\renewcommand{\arraystretch}{1.12}
\begin{tabular}{@{}>{\raggedright\arraybackslash}p{0.20\linewidth}>{\raggedright\arraybackslash}p{0.30\linewidth}>{\raggedright\arraybackslash}p{0.41\linewidth}@{}}
\toprule
\textbf{Dimension} & \textbf{Role} & \textbf{Common realizations} \\
\midrule
Input Assembly & Context entry preparation & Tool-result formatting; observation compression \\
\addlinespace[2pt]
History Maintenance & Historical context management & Compaction; summarization; persistent state \\
\addlinespace[2pt]
Context Orchestration & Context access and sharing across agents and subtasks & Instructions and tools for context access; subagent instructions and feedback design \\
\bottomrule
\end{tabular}
\end{table}

Formally, we represent a context policy as
\begin{equation}
P =
\left(
P^{\mathrm{in}},
P^{\mathrm{hist}},
P^{\mathrm{orch}}
\right)
\in \mathcal{P},
\label{eq:policy}
\end{equation}
where \(\mathcal{P}\) denotes the design space of harness context policies, and the three components correspond to the three dimensions above.
During evolution, \(P\) can be updated by modifying, combining, or introducing implementations along these dimensions.

\subsection{Evolution Objective}
The goal of context-policy evolution is to identify a policy that improves agent performance under a fixed base model \(M\): 
\begin{equation}  
P^* = \arg\max_{P\in\mathcal{P}} \mathbb{E}_{\tau\sim(M,H_P)}[R(\tau)], \label{eq:policy_optimization} \end{equation} 
where \(R(\tau)\) denotes the task-level reward.
The notation \(\tau\sim(M,H_P)\) denotes an execution under harness \(H_P\) with the context policy \(P\).

\section{Method}
\label{sec:method}

Under the evolution objective in Equation~\ref{eq:policy_optimization}, \method{} uses long-horizon execution trajectories to evolve a context policy while keeping the base model \(M\) fixed.
It addresses the two challenges identified above: determining whether a failure arises from how the harness manages information exposed to the model, and identifying which policy dimension should be evolved. 

During evolution, \method{} first reconstructs evidence about how information is handled across observed trajectories. 
It then uses this evidence across cases to attribute failures to context management and identify the corresponding policy dimensions. 
Supported attributions are used to update the context policy by modifying existing implementations, combining them, or introducing new ones along the identified dimensions.

We describe evidence reconstruction, failure attribution, and policy update in Sections \ref{sec:method_evidence}--\ref{sec:method_update}, respectively.
Within one evolution run, the acting agent, analysis agent, replay calls, and policy editor use the same fixed base model \(M\).

\begin{figure*}[t]
\centering
\includegraphics[width=\textwidth]{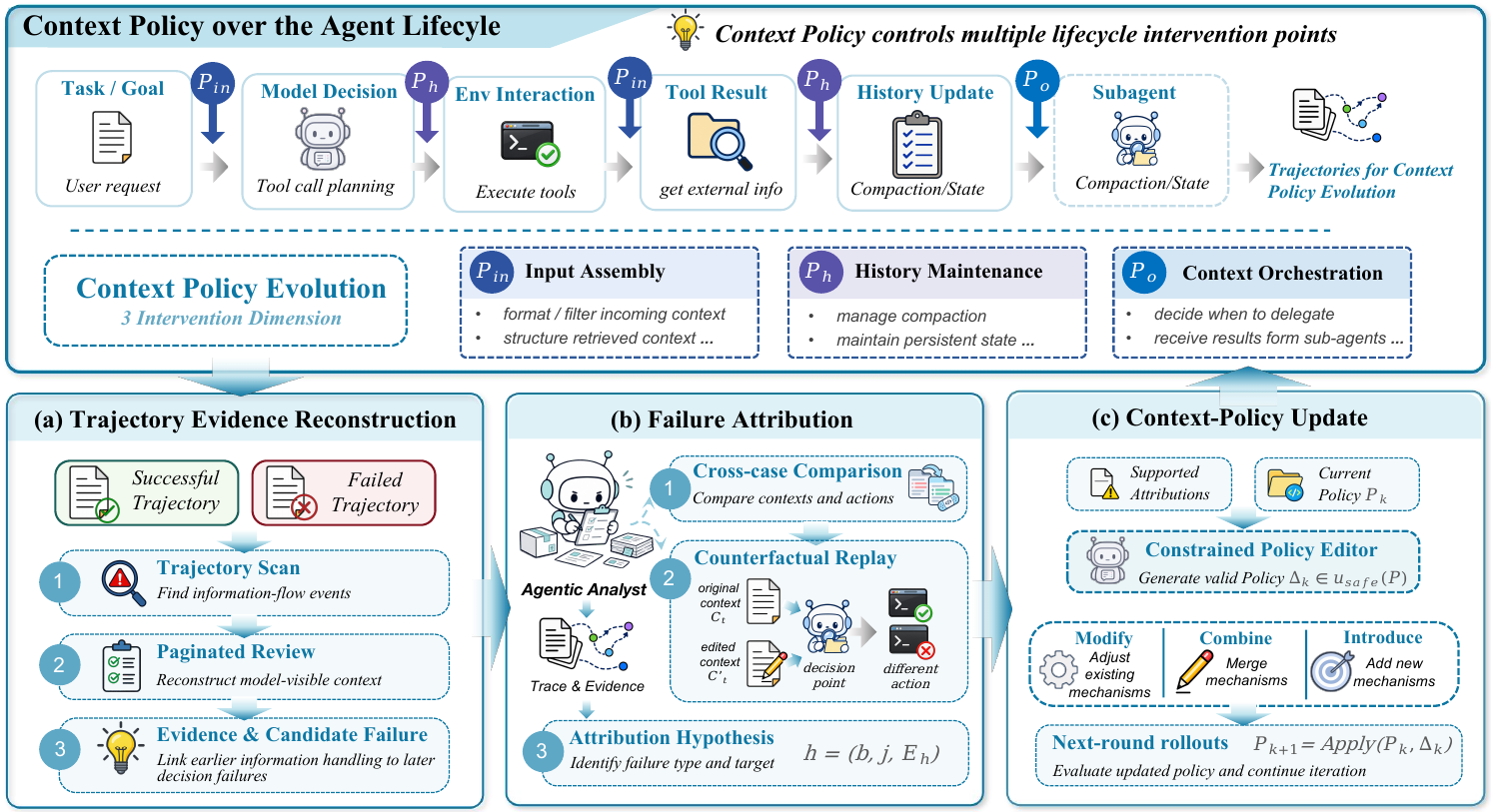}
\caption{\textbf{The overview of \method{}.} The context policy governs how information is presented, organized, and shared through the agent harness while the base model remains fixed. \method{} reconstructs trajectory evidence, attributes context-management failures to policy dimensions, and applies constrained updates for the next evolution round.}
\label{fig:method_overview}
\end{figure*}

\subsection{Trajectory Evidence Reconstruction}
\label{sec:method_evidence}
\emph{Evidence Reconstruction} establishes the evidence for subsequent failure attribution by recovering how information is exposed and carried through each trajectory. 
It combines a trajectory-wise information-flow scan with further review of the model-visible context.

\paragraph{Trajectory scan.}
Information relevant to a later decision may have appeared many steps earlier, and long trajectories may exceed a single analysis context. 
We therefore scan the full trajectory for a compact record of information-processing events across the execution, including when task-relevant information is presented, organized, or revisited, as well as task progress and repeated reads or executions.


\paragraph{Paginated review.}
The extracted record alone does not determine how information handling contributed to a failure. 
We therefore review the trajectory in successive segments, which we refer to as \emph{pages}, while reconstructing the model-visible context at relevant decision points. 
Throughout the review, the reviewer maintains a running \emph{note} that records task progress, unresolved events, and relevant context changes, updating it after each page. 
The note carries information across pages, allowing the review to relate earlier information-processing events to later decisions. 
The review identifies candidate context-management failures while distinguishing failures that may instead arise from the model, tools, or environment.


\subsection{Failure Attribution and Target Identification}
\label{sec:method_attribution}
Given the reconstructed trajectory evidence, \method{} determines whether a failure can be attributed to context management and, if so, which policy dimension should be adjusted. 
An analysis agent then examines candidate decision points, compares evidence across successful and failed trajectories, and uses counterfactual replay when the observed evidence does not resolve the attribution.

\paragraph{Counterfactual replay.}
Rerunning the full task is both costly and difficult to interpret, since later execution may diverge for reasons unrelated to the context change being tested. We instead hold the execution prefix \(\tau_{<t}\) fixed and modify only the model-visible context at decision \(t\),
$$
\widetilde{c}_{t}=\operatorname{Edit}(c_{t}).
$$
We then compare samples from the original and modified contexts:
\begin{equation}
a_t^{(r)}\sim M(\cdot\mid c_{t}),\qquad
\widetilde{a}_t^{(r)}\sim M(\cdot\mid \widetilde{c}_{t}),
\qquad r=1,\ldots,N.
\label{eq:decision_replay}
\end{equation}
This local comparison tests whether the proposed context-policy update improves the model's decision at the same execution point without side effects for subsequent actions.


\paragraph{Attribution hypothesis.}
The analysis produces a candidate attribution hypothesis together with the evidence supporting it. We represent an attribution hypothesis as
\begin{equation}
h=(b,j,\mathcal{E}_h),
\label{eq:attribution_hypothesis}
\end{equation}
where 
\(b\in\{\text{missing},\text{degraded},\text{buried},\text{polluted}\}\) 
characterizes the information failure,
\(j\in\{\mathrm{in},\mathrm{hist},\mathrm{orch}\}\) 
identifies the policy dimension to which the failure is attributed, and \(\mathcal{E}_h\) collects the supporting evidence. This evidence may include decision-point observations, cross-case comparisons, context-growth comparisons, replay results, and relevant counterexamples or alternative explanations. 

We retain an attribution only when the evidence distinguishes failed from successful executions and supports the proposed failure type and policy target.
Supported attributions are then passed to the policy update stage.



\subsection{Context-Policy Update}
\label{sec:method_update}

Given the supported attributions and the current context policy \(P_k\), the editor constructs a constrained update \(\Delta_k\). 
Consistent with the policy space defined in Section~\ref{sec:preliminaries}, the update may modify, combine, or introduce implementations along the dimensions, subject to fixed safety constraints. 
The context policy is updated as
\begin{equation}
P_{k+1}=\operatorname{Apply}(P_k,\Delta_k),\qquad
\Delta_k\in\mathcal{U}_{\mathrm{safe}}(P_k).
\label{eq:policy_update}
\end{equation}
Here \(\mathcal{U}_{\mathrm{safe}}(P_k)\) denotes the updates allowed by these constraints. If the available evidence does not support an update, \(\Delta_k=\emptyset\) and \(P_{k+1}=P_k\). The updated policy is then used to generate trajectories for the next evolution round, while held-out tasks are reserved for final evaluation of the selected policy \(P^*\). 

\section{Experiments and Results}
\label{sec:experiments}

\subsection{Experimental Setup}
\label{sec:exp_setup}

\newcommand{\scoregain}[1]{{\scriptsize\textcolor{green!45!black}{#1}}}
\newcommand{\scoreloss}[1]{{\scriptsize\textcolor{red!65!black}{#1}}}

\begin{table*}[t]
\label{tab:main_results}
\caption{\textbf{Performance of baseline harnesses and \method{} over two evolution iterations.}
Scores are reported on the evolution (Held-in), evaluation (Held-out), and full task sets.
Values in parentheses indicate percentage-point changes from the preceding baseline (Pi-agent for iter-1 and iter-1 for iter-2).
Bold and underline indicate the best and second-best scores, respectively.}
\centering
\footnotesize
\setlength{\tabcolsep}{1.5pt}
\renewcommand{\arraystretch}{1.03}
\resizebox{\textwidth}{!}{%
\begin{tabular}{@{}l*{9}{c}@{}}
\toprule
& \multicolumn{3}{c}{\scriptsize Long-Horizon TB} & \multicolumn{3}{c}{DeepSWE} & \multicolumn{3}{c}{BrowseComp-Plus} \\
\cmidrule(lr){2-4}\cmidrule(lr){5-7}\cmidrule(lr){8-10}
Harness / policy & {\scriptsize Held-in (\%)} & {\scriptsize Held-out (\%)} & {\scriptsize All (\%)} & {\scriptsize Held-in (\%)} & {\scriptsize Held-out (\%)} & {\scriptsize All (\%)} & {\scriptsize Held-in (\%)} & {\scriptsize Held-out (\%)} & {\scriptsize All (\%)} \\
\midrule
\multicolumn{10}{@{}l}{\textit{Harness (DeepSeek-V4-Flash)}} \\
\hspace{0.8em}OpenCode & 36.1 & 40.3 & 38.5 & \underline{62.5} & 66.2 & 64.6 & \textbf{92.1} & 77.6 & 83.9 \\
\hspace{0.8em}Codex & 41.8 & 40.1 & 40.8 & 45.8 & 63.1 & 55.8 & 84.2 & 84.5 & 84.4 \\
\hspace{0.8em}OpenClaw & 36.9 & \textbf{45.9} & 42.0 & 45.8 & 60.0 & 54.0 & 89.5 & \textbf{88.8} & \textbf{89.1} \\
\hspace{0.8em}\textbf{Pi-agent (base)} & 43.3 & 40.4 & 41.7 & 54.2 & 72.3 & 64.6 & 81.6 & 82.7 & 82.2 \\
\cmidrule(lr){1-10}
\addlinespace[2pt]
\multicolumn{10}{@{}l}{\textit{\methodname{} (DeepSeek-V4-Flash)}} \\
\hspace{0.8em}\textbf{iter-1} & \textbf{47.0}\,\scoregain{(+3.7)} & 43.1\,\scoregain{(+2.7)} & \underline{44.8}\,\scoregain{(+3.1)} & 60.4\,\scoregain{(+6.3)} & \underline{75.4}\,\scoregain{(+3.1)} & \underline{69.0}\,\scoregain{(+4.4)} & \underline{90.8}\,\scoregain{(+9.2)} & 81.6\,\scoreloss{(-1.0)} & 85.6\,\scoregain{(+3.4)} \\
\addlinespace[2pt]
\hspace{0.8em}\textbf{iter-2} & \underline{45.7}\,\scoreloss{(-1.3)} & \underline{45.0}\,\scoregain{(+1.9)} & \textbf{45.3}\,\scoregain{(+0.6)} & \textbf{66.7}\,\scoregain{(+6.3)} & \textbf{75.4}\,\scoregain{(+0.0)} & \textbf{71.7}\,\scoregain{(+2.7)} & \textbf{92.1}\,\scoregain{(+1.3)} & \underline{85.7}\,\scoregain{(+4.1)} & \underline{88.5}\,\scoregain{(+2.9)} \\
\midrule
\multicolumn{10}{@{}l}{\textit{Harness (GPT-5.6-Luna)}} \\
\hspace{0.8em}OpenCode & 32.0 & 32.3 & 32.1 & \underline{22.9} & \underline{21.5} & \underline{22.1} & 35.5 & 34.7 & 35.1 \\
\hspace{0.8em}Codex & \textbf{36.3} & \textbf{38.6} & \textbf{37.6} & \textbf{43.8} & \textbf{44.6} & \textbf{44.3} & \textbf{64.0} & \textbf{57.1} & \textbf{60.1} \\
\hspace{0.8em}OpenClaw & 20.5 & 17.4 & 18.8 & 10.4 & 10.8 & 10.6 & 22.4 & 16.3 & 19.0 \\
\hspace{0.8em}\textbf{Pi-agent (base)} & 31.8 & 29.2 & 30.3 & 6.3 & 18.5 & 13.3 & 35.5 & 39.8 & 37.9 \\
\cmidrule(lr){1-10}
\addlinespace[2pt]
\multicolumn{10}{@{}l}{\textit{\methodname{} (GPT-5.6-Luna)}} \\
\hspace{0.8em}\textbf{iter-1} & 27.6\,\scoreloss{(-4.2)} & 33.7\,\scoregain{(+4.5)} & 31.0\,\scoregain{(+0.7)} & 6.3\,\scoregain{(+0.0)} & 15.4\,\scoreloss{(-3.1)} & 11.5\,\scoreloss{(-1.8)} & \underline{46.1}\,\scoregain{(+10.5)} & 33.7\,\scoreloss{(-6.1)} & 39.1\,\scoregain{(+1.1)} \\
\hspace{0.8em}\textbf{iter-2} & \underline{33.7}\,\scoregain{(+6.1)} & \underline{34.8}\,\scoregain{(+1.0)} & \underline{34.3}\,\scoregain{(+3.3)} & 14.6\,\scoregain{(+8.3)} & 13.8\,\scoreloss{(-1.5)} & 14.2\,\scoregain{(+2.7)} & 43.4\,\scoreloss{(-2.6)} & \underline{40.8}\,\scoregain{(+7.1)} & \underline{42.0}\,\scoregain{(+2.9)} \\
\bottomrule
\end{tabular}
}%
\end{table*}

\paragraph{Tasks and evaluation.}
We evaluate \method{} in three long-horizon benchmark settings.
Long-Horizon-Terminal-Bench (LHTB) \citep{li2026longhorizonterminal} contains 46 terminal workflows, DeepSWE \citep{huang2026deepswe} contains 113 repository-level software-engineering tasks, and BrowseComp-Plus \citep{chen2026browsecompplus} uses a hard, search-intensive subset of 174 questions.
LHTB uses a dense task score, while DeepSWE and BrowseComp-Plus use binary task scores from verifiers.
For each benchmark, we report results on the full task set and separately on the evolution set (Held-in) and reserved evaluation set (Held-out).
Benchmark distributions and details, along with the construction and motivation of the BrowseComp-Plus subset are detailed in Appendix~\ref{app:benchmark_protocol}.

\paragraph{Baselines and evolution settings.}
We evaluate four agent harness baselines, i.e., OpenCode, Pi-agent, Codex, and OpenClaw, with two different models: DeepSeek-V4-Flash-0731 and GPT-5.6-Luna.
We implement \method{} in the open-source Pi-agent harness and run two successive evolution iterations to assess whether performance continues to improve beyond the first iteration. 
Within each benchmark-model setting, we keep the base model used for failure attribution and target identification fixed across evolution iterations. 
The evolution pipeline uses only held-in trajectories, and each evolved policy and its corresponding baseline are evaluated on the same task split under identical execution settings. 
Model versions, inference parameters, and harness configurations are provided in Appendix~\ref{app:implementation_details}.

\subsection{Main Results}
\label{sec:exp_main_results}

Table~\ref{tab:main_results} compares the initial baseline and the \method{} policies with established harnesses.

\textbf{\method{} is effective across different base models and task settings, while remaining highly competitive with prominent harnesses.} After evolution, ContextEvo achieves the best full-benchmark scores on LHTB and DeepSWE under DeepSeek-V4-Flash, reaching 45.3\% and 71.7\%, respectively, and scores 88.5\% on BrowseComp-Plus, only 0.6\% below OpenClaw. Under GPT-5.6-Luna, the evolved policy also improves over the initial baseline on all three benchmarks and ranks second only to the Luna-native Codex harness on LHTB and BrowseComp-Plus, with scores of 34.3\% and 42.0\%, respectively. These results show that \method{} is broadly effective across models and task settings while achieving performance comparable to established and strong harnesses.

\textbf{\method{} enables sustained gains across successive evolution iterations.}
On the full task sets, iter-2 improves over iter-1 in all six benchmark--model settings.
Under DeepSeek-V4-Flash, the scores increase from 44.8\% to 45.3\% on LHTB, from 69.0\% to 71.7\% on DeepSWE, and from 85.6\% to 88.5\% on BrowseComp-Plus.
The same trend holds under GPT-5.6-Luna, where the scores increase from 31.0\% to 34.3\%, from 11.5\% to 14.2\%, and from 39.1\% to 42.0\%, respectively.
These consistent second-round gains indicate that \method{} can continue refining the context policy beyond the initial update rather than producing only a one-time improvement.

\section{Analysis}
\label{sec:analysis}

\subsection{Beyond Skill Evolution in Long-Horizon Tasks}



\begin{table}[!ht]
    \centering
    \caption{\textbf{Context-policy vs. skill evolution on LHTB-46.}
    Reward and mean token usage per run are reported for different settings.}
    \label{tab:lhtb_skill_context}

    \footnotesize
    \setlength{\tabcolsep}{5pt}

    \begin{tabular}{@{}lcc@{}}
        \toprule
        Agent & Reward $\uparrow$ & Tokens (M) $\downarrow$ \\
        \midrule
        Baseline (Pi)        & 0.417          & 8.57 \\
        Skill-only           & 0.370          & 9.61 \\
        \method{}            & \textbf{0.448} & \textbf{7.52} \\
        \method{} + skill    & 0.419          & 9.69 \\
        \bottomrule
    \end{tabular}
    \vspace{-13pt}   
\end{table}

\textbf{Context-policy evolution outperforms skill-only evolution in the long-horizon setting.}
Following recent trajectory-driven harness evolution \citep{zhang2026selfharness}, we derive procedural skills from the same held-in LHTB tasks used by \method{}.
Across 46 tasks, skill-only reward falls from 0.417 to 0.370, while mean tokens rise from 8.57M to 9.61M (Table~\ref{tab:lhtb_skill_context}).
\method{} reaches 0.448 with 7.52M tokens; adding skills to it yields 0.419 with 9.69M.

The trajectories reveal three limits of skill evolution.
\textbf{(1) Invocation gaps.} In long-horizon tasks, complex contexts make skills harder to use at the right decision points: only 20 of 44 inspectable skill-only trajectories read a skill body.
\textbf{(2) Local goal drift.} Even when used, skill guidance can divert long-horizon execution toward a local objective. In the DuckDB case, the agent uses the skill to pass local checks, but its task objective shifts and it does not continue refining the final patch.
\textbf{(3) Partial workflow coverage.} A skill evolved from long-horizon trajectories may cover only part of a larger workflow. In the MODFLOW audit, the agent follows a generic benchmark-loop skill and completes the artifacts requested by the skill, but the final submission remains incomplete relative to the requirements of the full workflow.
Together, these patterns motivate a full-lifecycle context policy that keeps task evidence available from initial observations through final delivery.
Additional skill-only cases and analyses are provided in Appendix~\ref{app:analysis_details}.

\subsection{Context Management Baselines}
\label{sec:context_management_baselines}

\begin{wraptable}[12]{l}{0.49\textwidth}
\vspace{-1.5\baselineskip}
\caption{Performance comparison across context-management methods.}
  \centering
  \scriptsize
  \setlength{\tabcolsep}{2pt}
  \resizebox{\linewidth}{!}{%
  \begin{tabular}{@{}lccc@{}}
  \toprule
  Method & DeepSWE & LHTB & \shortstack{BrowseComp-Plus} \\
  \midrule
  Pi-agent base & 0.133 & 0.303 & 0.379 \\
  \addlinespace[2pt]
  \multicolumn{4}{@{}l}{\textit{Static context management}} \\
  ReSum & 0.062 & 0.266 & 0.385 \\
  ACON & 0.124 & 0.262 & 0.374 \\
  \addlinespace[2pt]
  \multicolumn{4}{@{}l}{\textit{Dynamic context management}} \\
  TACO & 0.115 & 0.281 & 0.374 \\
  \addlinespace[2pt]
  \textbf{\method{}} & \textbf{0.142} & \textbf{0.343} & \textbf{0.420} \\
  \bottomrule
  \end{tabular}}%
 \label{tab:context_management_baselines}
  \end{wraptable}

Following the skill-evolution comparison, we compare static compression (ReSum and ACON) \citep{wu2026resum,kang2025acon}, local compression evolution (TACO) \citep{ren2026taco}, and full context-policy evolution (\method{}). All conditions use GPT-5.6-Luna under matched Pi-agent protocols. This isolates the full-policy effect.

\method{} achieves the highest score on all three benchmarks: 0.142 on DeepSWE, 0.343 on LHTB, and 0.420 on BrowseComp-Plus. This indicates that full-policy evolution better addresses long-horizon context failures than static compression or local adjustment alone. Additional cases appear in Appendix~\ref{app:analysis_details}.

\subsection{Long-Horizon Environment-Specific Policy Evolution}
\label{sec:environment_policy_evolution}

\begin{figure*}[!ht]
\centering
\includegraphics[width=\textwidth]{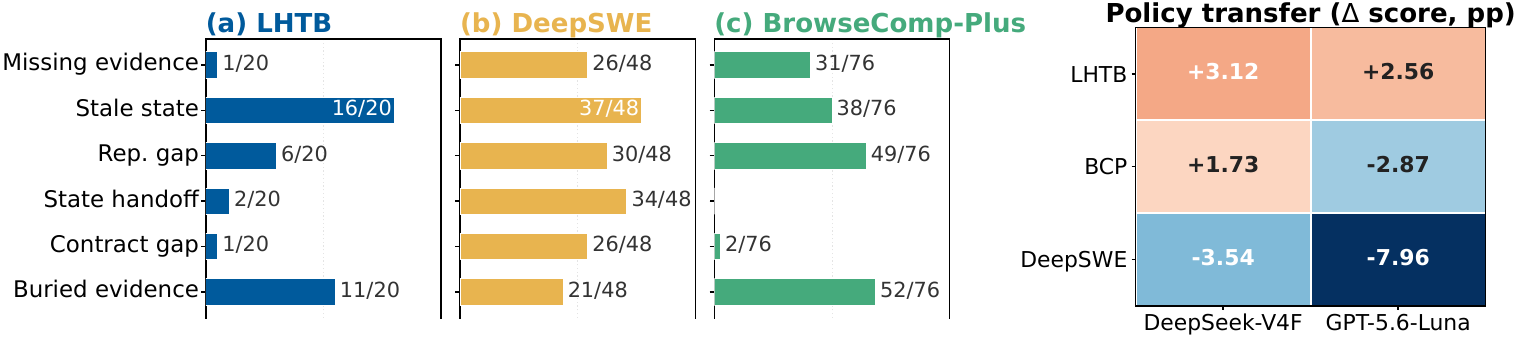}
\caption{\textbf{Environment-specific context problems and policy transfer.} Bars show the share of audited cases with each multi-select context problem; labels report counts. The heatmap shows signed score changes (percentage points) from applying the LHTB-evolved policy under DeepSeek and Luna.}
\label{fig:policy_case_problem_prevalence}
\end{figure*}

Across the three benchmarks, task environments impose different context pressures.
Figure~\ref{fig:policy_case_problem_prevalence} shows that \method{} learns different context-policy realizations from these pressures: selective observation handling and persistent execution state for LHTB, specification and verification continuity for DeepSWE, and recoverable evidence organization for BrowseComp-Plus.
Thus, \method{} adapts the context policy to each task environment rather than applying one fixed strategy across tasks.
The appendix provides additional case details and analysis, including examples where policy evolution introduces context-management mechanisms absent from the initial harness.

These learned policies transfer across models within the same environment, but not reliably across environments.
The LHTB policy improves both DeepSeek and Luna, whereas applying the frozen LHTB policy to DeepSWE decreases performance for both models and produces mixed effects on BrowseComp-Plus (Figure~\ref{fig:policy_case_problem_prevalence}).
This shows that policies can be reused across models when the context pressure is shared, while environments with substantially different pressures require renewed policy evolution.

\subsection{Component Ablation}
\label{sec:ablation}

We next evaluate the three-stage evolution pipeline on LHTB by comparing the native Pi-agent pipeline, the complete ContextEvo pipeline, and leave-one-operator-out variants under the same DeepSeek-V4-Flash-0731 protocol.
Table~\ref{tab:component_ablation} reports results on the held-in and held-out splits.

\begin{table}[t]
\caption{LHTB-46 component ablations with DeepSeek-V4-Flash-0731 under the matched Pi-agent protocol; $\Delta$ denotes the all-task difference from Full.}
\centering
\small
\setlength{\tabcolsep}{3.5pt}
\begin{tabular}{lrrrr}
\toprule
Variant & Held-in & Held-out & All & $\Delta$ vs. Full \\
\midrule
Native Pi-agent & 43.30\% & 40.39\% & 41.66\% & $-3.13$ pp \\
Full ContextEvo & 46.95\% & 43.12\% & \textbf{44.78\%} & -- \\
w/o Paginated Trajectory Review & 41.01\% & 26.19\% & 32.64\% & $-12.15$ pp \\
w/o Cross-Case Aggregation & 40.46\% & 44.05\% & 42.49\% & $-2.29$ pp \\
w/o Counterfactual Replay & 48.57\% & 35.29\% & 41.06\% & $-3.72$ pp \\
w/o Context-Policy Update & 32.57\% & 42.78\% & 38.34\% & $-6.44$ pp \\
\bottomrule
\end{tabular}
\label{tab:component_ablation}
\end{table}

\textbf{The complete evolution pipeline achieves the strongest overall performance.} Full reaches 44.78\%, improving over Native at 41.66\% by 3.12 percentage points.
\textbf{Evidence Reconstruction is the most consequential stage.} Removing Paginated Trajectory Review lowers the overall score to 32.64\% ($-12.15$ pp) and the held-out score to 26.19\% ($-16.93$ pp).
\textbf{Failure Attribution and Context-Policy Update provide complementary gains.} Removing Cross-Case Aggregation, Counterfactual Replay, or Context-Policy Update lowers the overall score by 2.29, 3.72, and 6.44 percentage points, respectively; the replay ablation is especially damaging on held-out tasks, reducing the score from 43.12\% to 35.29\%. Together, these results show that the strongest performance comes from the complete evolution loop.

\textbf{Cross-harness validation.} To check that the gains are not confined to Pi-agent, we also evaluate ContextEvo on OpenCode and OpenClaw with GPT-5.6-Luna on BrowseComp-Plus hard174. ContextEvo improves OpenCode from 35.06\% (61/174) to 40.80\% (71/174), a gain of 10 tasks (+5.74 pp), and OpenClaw from 18.97\% (33/174) to 45.40\% (79/174), a gain of 46 tasks (+26.44 pp). The matched protocol and full comparison are reported in Appendix~\ref{app:cross_harness_validation}.

\section{Conclusion}
\label{sec:conclusion}

We presented ContextEvo, a framework that evolves a context policy for long-horizon agent execution while keeping the base model fixed within each benchmark--model run.
The policy spans three dimensions---Input Assembly, History Maintenance, and Context Orchestration---and is updated through evidence reconstruction, failure attribution and target identification, and constrained policy update.

Across the reported benchmark and model settings, the selected policies improve several full-benchmark point estimates over the Pi-agent baseline, while held-out changes vary by setting.
Case-level analysis indicates that the reviewed updates emphasize different information demands in terminal workflows, repository engineering, and multi-hop search.

The evidence also defines the limits of the method's claims.
Cross-case evidence supports an attribution hypothesis, bootstrap evidence can justify a bounded update, and decision-point replay measures only local next-action sensitivity.
These forms of evidence do not by themselves establish a task-level causal reward contribution for an individual policy mechanism.

\section*{AI use statement}
In this work, we used generative AI tools to assist with result interpretation, literature organization, and the drafting, editing, and formatting of text, tables, and figures. The authors reviewed and verified all AI-assisted outputs, including claims, citations, analyses, and presentation, and take responsibility for the final content of this work.

\bibliographystyle{iclr2027_conference}
\bibliography{iclr2027_conference}
\clearpage
\appendix
\section{Benchmark and Experimental Protocol}
\label{app:benchmark_protocol}

\subsection{Benchmark settings}
\label{app:benchmark_settings}

\begin{figure*}[t]
\centering
\includegraphics[width=\textwidth]{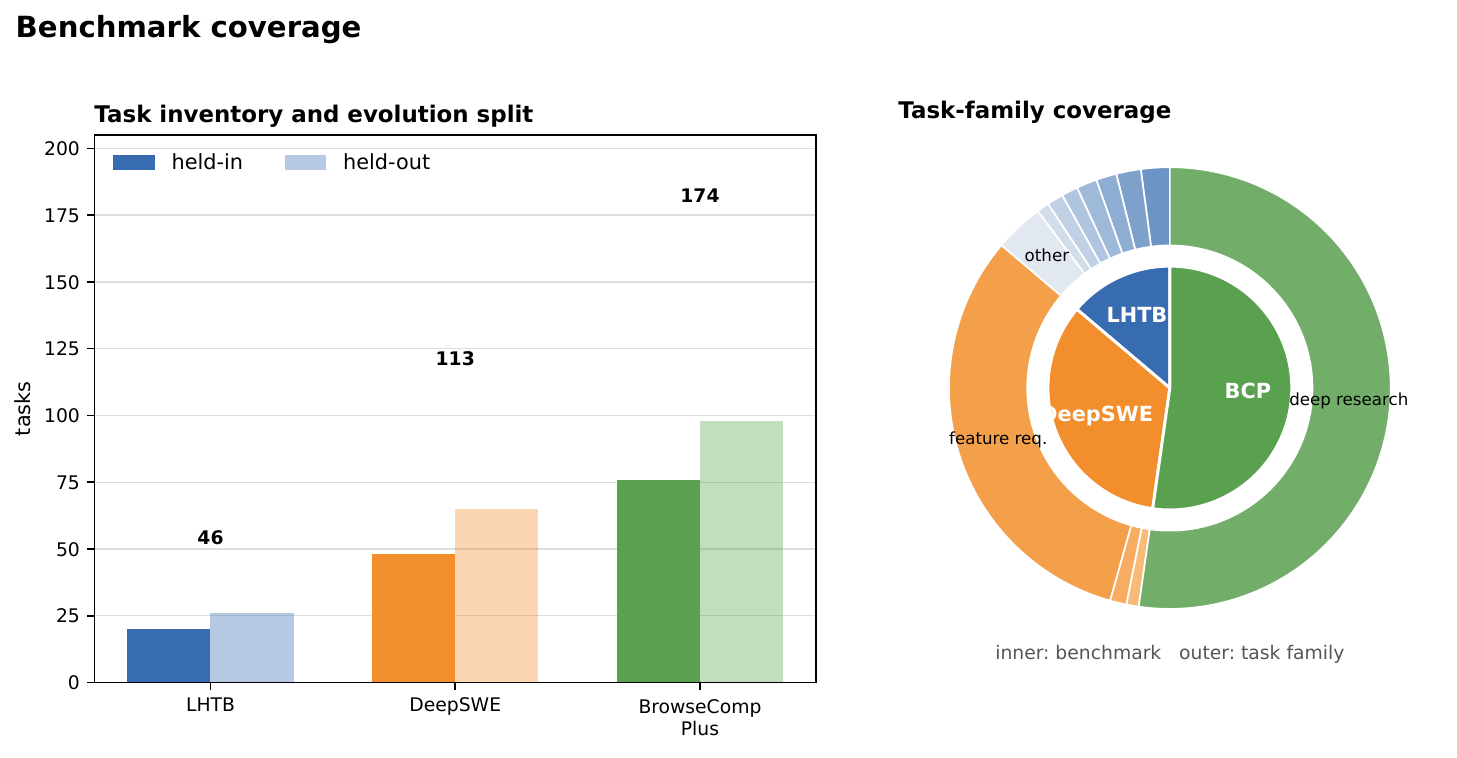}
\caption{\textbf{Benchmark coverage.} The left panel reports the full inventory and the held-in/held-out split. The sunburst shows the benchmark inventory in the inner ring and task-family composition in the outer ring. LHTB shows its seven largest categories plus an ``other'' bucket, DeepSWE shows request types, and BrowseComp-Plus is a uniform multi-hop deep-research setting.}
\label{fig:benchmark_profile}
\end{figure*}

\begin{figure*}[t]
\centering
\includegraphics[width=\textwidth]{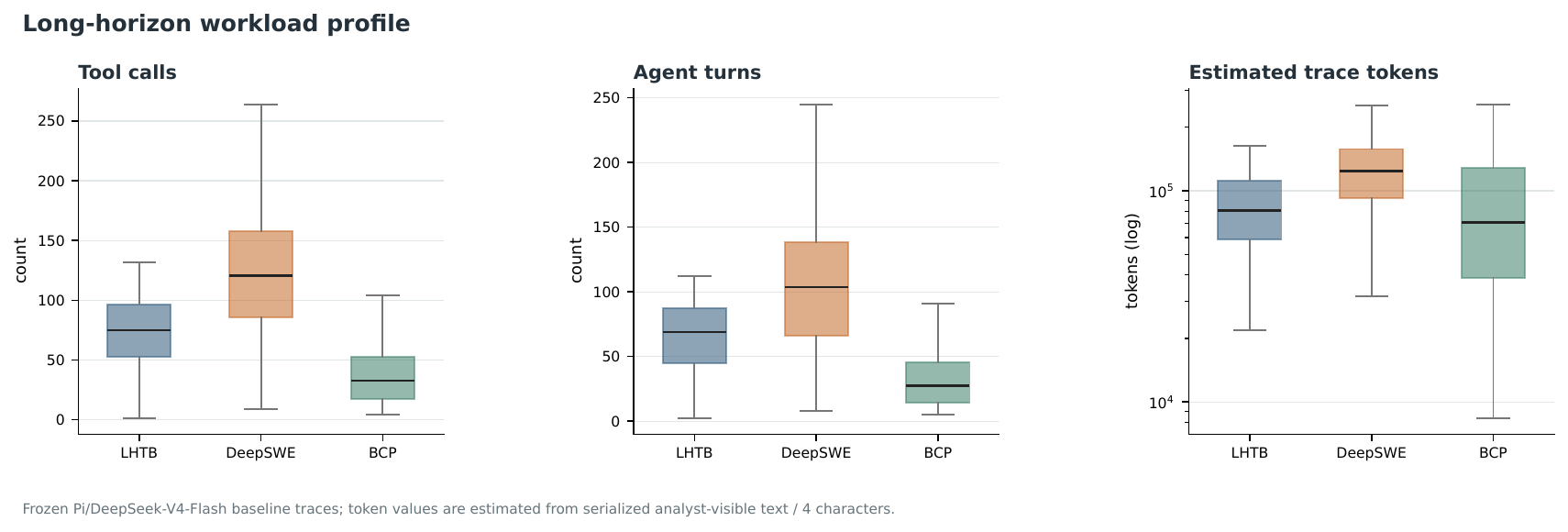}
\caption{\textbf{Long-horizon workload profile.} The three panels show the distributions of tool calls, agent turns, and estimated trajectory text volume in frozen Pi/DeepSeek-V4-Flash baseline traces. The token measure is an audit estimate from serialized analyst-visible text rather than provider billing tokens; the corresponding median and interquartile values are listed in Table~\ref{tab:workload_profile}.}
\label{fig:workload_profile}
\end{figure*}

\paragraph{Benchmark profiles.}
We evaluate three task settings with different sources of long-horizon context. Long-Horizon-Terminal-Bench (LHTB) contains terminal workflows that span software engineering, games, tool-use, multimodal work, and scientific or systems tasks. DeepSWE contains repository-level software-engineering requests, dominated by feature requests with a smaller set of bug fixes and enhancements. BrowseComp-Plus contains multi-hop deep-research questions that require repeated retrieval and evidence synthesis. Table~\ref{tab:benchmark_inventory} gives the task counts and fixed evolution split; Figure~\ref{fig:benchmark_profile} shows the task-family coverage, while Figure~\ref{fig:workload_profile} and Table~\ref{tab:workload_profile} report workload statistics measured from the baseline trajectories.

\begin{table}[t]
\caption{Benchmark inventories and fixed evolution splits. Held-in tasks provide trajectories for evidence reconstruction, attribution, replay, and policy updating; held-out tasks are reserved for evaluation.}
\label{tab:benchmark_inventory}
\centering
\small
\setlength{\tabcolsep}{5pt}
\begin{tabular}{@{}l l r r r@{}}
\toprule
Benchmark & Scenario & All & Held-in & Held-out \\
\midrule
LHTB & terminal workflows & 46 & 20 & 26 \\
DeepSWE & repository software engineering & 113 & 48 & 65 \\
BrowseComp-Plus & multi-hop deep research & 174 & 76 & 98 \\
\bottomrule
\end{tabular}
\end{table}

\begin{table*}[t]
\caption{Workload statistics from frozen Pi/DeepSeek-V4-Flash baseline trajectories. Entries are median $[\mathrm{p25},\mathrm{p75}]$; trace tokens estimate analyst-visible text and are not billing tokens.}
\label{tab:workload_profile}
\centering
\small
\setlength{\tabcolsep}{3pt}
\begin{tabular}{@{}l r c c c@{}}
\toprule
Benchmark & Trajectories & Tool calls & Agent turns & Estimated trace tokens \\
\midrule
LHTB & 44 & $75.0\;[52.5,\,96.5]$ & $69.0\;[45.0,\,87.3]$ & $80{,}712\;[58{,}862,\,111{,}707]$ \\
DeepSWE & 112 & $120.5\;[85.8,\,158.0]$ & $103.5\;[65.8,\,138.5]$ & $124{,}205\;[92{,}726,\,157{,}711]$ \\
BrowseComp-Plus & 174 & $32.5\;[17.3,\,52.8]$ & $27.5\;[14.0,\,45.8]$ & $70{,}866\;[38{,}626,\,128{,}303]$ \\
\bottomrule
\end{tabular}
\end{table*}

\paragraph{BrowseComp-Plus hard subset.}
The official BrowseComp-Plus release has 830 questions.
We use a 174-question hard subset because the full benchmark produces a ceiling effect for the current frontier model used in our main comparison.
We first grade the publicly released o3 and GPT-5 reference answers with the benchmark judge, retain questions that both models miss, and then require at least 30 o3 searches.
This yields questions that are both difficult for frontier references and search-intensive.
We apply \texttt{random.Random(15)} to the sorted task names and assign the first 76 questions to held-in and the remaining 98 to held-out.
The held-out questions are never used for trajectory scan, paginated trajectory review, failure attribution, replay, or policy updating.
The complete construction and offline scoring procedure, including answer extraction and infrastructure-error handling, are documented here so that the main text need only state the resulting subset and split.

\begin{table}[t]
\caption{BrowseComp-Plus hard-subset construction.}
\label{tab:bcp_subset}
\centering
\small
\setlength{\tabcolsep}{5pt}
\begin{tabular}{@{}l l@{}}
\toprule
Selection stage & Criterion \\
\midrule
Reference difficulty & o3 and GPT-5 reference answers are both judged incorrect \\
Long-horizon filter & o3 issues at least 30 searches \\
Resulting subset & 174 questions from the 830-question release \\
Evolution split & seed 15; 76 held-in / 98 held-out \\
\bottomrule
\end{tabular}
\end{table}

\subsection{Evaluation configuration}
\label{app:evaluation_configuration}
\label{app:implementation_details}

\paragraph{Model configurations.}
\label{app:model_configurations}

We use the same model configurations across the three benchmarks. Table~\ref{tab:model_configurations} records the model release, provider identifier, reasoning setting, context limit, output limit, and generation parameters.

\begin{table}[t]
\caption{Model configurations used in the experiments.}
\label{tab:model_configurations}
\centering
\small
\setlength{\tabcolsep}{4pt}
\begin{tabular}{@{}>{\raggedright\arraybackslash}p{0.28\linewidth}>{\raggedright\arraybackslash}p{0.62\linewidth}@{}}
\toprule
Setting & Configuration \\
\midrule
DeepSeek-V4-Flash-0731 & \texttt{openai/deepseek-v4-flash}; reasoning enabled; context 262{,}144; max output 32{,}768 \\
GPT-5.6-Luna & \texttt{openai/gpt-5.6-luna}; reasoning enabled; context 262{,}144; max output 32{,}768 \\
Generation parameters & Temperature and top-$p$ use the provider defaults; no additional sampling override \\
\bottomrule
\end{tabular}
\end{table}

\paragraph{Harness configurations.}
\label{app:harness_configurations}

Table~\ref{tab:harness_configurations} records the execution settings for the four harnesses. The task timeout is 5{,}400 seconds for the reported runs; setup limits and trial parallelism follow the benchmark configurations. Each task is attempted once, and the baseline and self-evolved policy use the same harness settings within a comparison.

\begin{table}[t]
\caption{Harness execution settings. Concurrent trial counts are LHTB/DeepSWE/BCP.}
\label{tab:harness_configurations}
\centering
\small
\setlength{\tabcolsep}{4pt}
\begin{tabular}{@{}>{\raggedright\arraybackslash}p{0.24\linewidth}>{\raggedright\arraybackslash}p{0.66\linewidth}@{}}
\toprule
Harness & Execution configuration \\
\midrule
OpenCode & v1.15.13-20260715; OpenAI-compatible protocol; context/output 262{,}144/32{,}768; task timeout 5{,}400~s; 46/32/64 concurrent trials \\
Pi-agent & v0.80.6; \texttt{openai-responses}; context/output 262{,}144/32{,}768; task timeout 5{,}400~s; 24/32/64 concurrent trials \\
Codex & Frozen comparison release; \texttt{openai-chat}; task timeout 5{,}400~s; 46/32/64 concurrent trials \\
OpenClaw & v2026.6.1; \texttt{openai-completions}; context/output 262{,}144/32{,}768; task timeout 5{,}400~s; 46/32/64 concurrent trials \\
\bottomrule
\end{tabular}
\end{table}

\paragraph{Scoring and result handling.}
\label{app:score_accounting}

Let $U$ denote one of the fixed held-in, held-out, or full task subsets. For LHTB, each task receives a partial reward $r_t^{\mathrm{LHTB}}\in[0,1]$, and the reported score is
\begin{equation}
S_{\mathrm{LHTB}}(U)=\frac{1}{|U|}\sum_{t\in U}r_t^{\mathrm{LHTB}}.
\label{eq:lhtb_score}
\end{equation}

For DeepSWE, let $y_t^{\mathrm{DeepSWE}}\in\{0,1\}$ indicate whether the repository-level task is solved according to the benchmark evaluator. The score is
\begin{equation}
S_{\mathrm{DeepSWE}}(U)=\frac{1}{|U|}\sum_{t\in U}y_t^{\mathrm{DeepSWE}}.
\label{eq:deepswe_score}
\end{equation}

For BrowseComp-Plus, let $y_t^{\mathrm{BCP}}\in\{0,1\}$ indicate whether the submitted answer is judged correct. The score is
\begin{equation}
S_{\mathrm{BCP}}(U)=\frac{1}{|U|}\sum_{t\in U}y_t^{\mathrm{BCP}}.
\label{eq:bcp_score}
\end{equation}

We report these means as percentages. An agent timeout or an execution failure attributable to the agent receives a score of zero, with the timeout limit determined by the corresponding benchmark configuration. A trial is rerun only when the failure is confirmed to be caused by the evaluation infrastructure; the confirmed rerun result replaces the failed run before the benchmark score is computed.

\section{Method Details}
\label{app:method_details}


\subsection{Context-policy realization}
\label{app:contextprogram_realization}

The editor operates over eight context intervention points exposed at the harness--policy boundary. At runtime, the fixed harness adapter maps native execution events to the corresponding lifecycle points and invokes the ContextProgram where applicable. Decisions from these points jointly determine the context exposed to the agent. The table lists the current realization and illustrative policy modifications at each point; an evolution step may leave a point unchanged, span multiple points, or introduce a mechanism that is not present in the current realization.
The ContextProgram is the executable realization of the abstract context policy \(P\) inside the fixed harness \(H\).
The eight intervention points are concrete implementation surfaces grouped under the three policy dimensions; an attribution target \(j\) identifies a dimension and may therefore map to one or more intervention points.

\begin{table*}[t]
\caption{\textbf{Context-policy intervention points.} Current realizations and representative policy modifications.}
\label{tab:context_intervention_points}
\centering
\small
\setlength{\tabcolsep}{2pt}
\begin{tabular}{@{}p{0.03\textwidth}>{\raggedright\arraybackslash}p{0.15\textwidth}>{\raggedright\arraybackslash}p{0.20\textwidth}>{\raggedright\arraybackslash}p{0.20\textwidth}>{\raggedright\arraybackslash}p{0.32\textwidth}@{}}
\toprule
& Policy interface & Lifecycle intervention point & Current realization & Examples of possible policy modifications \\
\midrule
1
& Input Assembly
& Observation entry
& \shortstack[l]{\texttt{observation\_}\\[-1pt]\texttt{projector}}
& Change whether a newly received observation is retained in its original form or represented by a shorter, source-linked form. \\

2
& History Maintenance
& Compaction planning
& \shortstack[l]{\texttt{compaction\_}\\[-1pt]\texttt{planner}}
& Change when historical material becomes eligible for transformation and which complete dependency groups may be selected. \\

3
& History Maintenance
& Historical representation update
& \texttt{summary\_writer}
& Change which facts, constraints, validation results, failed approaches, and source references are preserved in a compact representation. \\

4
& History Maintenance
& Evidence memory update
& \texttt{memory\_writer}
& Change how durable evidence records are added, replaced, or removed and how their provenance is maintained. \\

5
& History Maintenance
& Evidence recovery
& \shortstack[l]{\texttt{memory\_}\\[-1pt]\texttt{retriever}}
& Change when prior evidence is requested and which narrowly relevant records are recovered. \\

6
& Input Assembly
& Request assembly
& \shortstack[l]{\texttt{context\_}\\[-1pt]\texttt{assembler}}
& Change how active history, summaries, recovered evidence, and inherited context are selected and ordered. \\

7
& Context Orchestration
& Parent-to-child context dispatch
& \shortstack[l]{\texttt{child\_context\_}\\[-1pt]\texttt{builder}}
& Change which parent evidence is authorized for an already specified child task. \\

8
& Context Orchestration
& Child-to-parent result return
& \shortstack[l]{\texttt{child\_result\_}\\[-1pt]\texttt{reducer}}
& Change how child findings, evidence references, uncertainty, and unresolved issues are represented to the parent. \\
\bottomrule
\end{tabular}
\end{table*}

\paragraph{Lifecycle interface implementation.}
\label{app:lifecycle_interface_implementation}

In our implementation, the eight intervention points are exposed through a versioned ContextProgram connected to the agent by a fixed harness adapter. The adapter maps native execution events to the applicable lifecycle points, provides the state available at each point, and applies the resulting context decisions before subsequent execution. Observation handling may preserve an original result or produce a source-linked representation; history handling may transform, summarize, update, or recover earlier evidence; request assembly determines the next model-visible view; and context orchestration controls which evidence crosses execution scopes and how returned results are represented. A point may remain unchanged for a given update, and a policy change may combine behavior across multiple points or introduce a new mechanism.
Thus, the intervention-point table instantiates the three dimensions rather than replacing them.
\(P^{\mathrm{in}}\) governs entry and request assembly, \(P^{\mathrm{hist}}\) governs retention and recovery, and \(P^{\mathrm{orch}}\) governs cross-scope access and return.

At runtime, the adapter does not execute the eight intervention points as a fixed sequence on every turn. It invokes the points whose lifecycle events are present. A newly produced observation first reaches the observation-entry point; when the accumulated history requires maintenance, the applicable planning, representation-update, memory-update, or recovery operations are invoked before the next model-visible request is assembled. Request assembly then combines the active history with any transformed or recovered evidence. The parent-to-child and child-to-parent points are invoked at their corresponding scope boundaries, so their decisions govern evidence entering a child and evidence returned to its parent. Each component receives the state exposed at its invocation point and returns a context decision that the adapter applies to subsequent execution. Consequently, a policy update may affect one event, several events, or introduce a new intervention without changing the harness's overall execution protocol.


\subsection{Context-policy evolution}
\label{app:context_policy_evolution}


\paragraph{Evolution data boundary.}
For each benchmark and base-model block, an evolution run reads all trajectories in the fixed held-in split for that setting.
Held-out tasks are excluded from trajectory scan, paginated trajectory review, failure attribution, replay, policy-update inputs, and candidate selection until the policy is frozen for evaluation.
The baseline and the selected policy then use the same task split, model, harness, and runtime protocol.
The reported v1 policy starts from the Pi-agent baseline; where v2 is reported, it denotes the next selected policy in the corresponding lineage in Table~\ref{tab:main_results}.
An evolution round may return an empty changeset when the evidence does not support an update, and only the resulting frozen policy \(P^*\) is evaluated on held-out tasks.
Trajectories from another benchmark or base-model block are not pooled into the same evolution run.

\paragraph{Deterministic case-card extraction.}
Stage~1a runs \texttt{tools/case\_cards.py} once per trial and reads only ledger requests that carry tool schemas; out-of-band generation calls are not counted as agent decisions.
The extractor skips benchmark bookkeeping directories, optionally restricts trials with the fixed split file, and can replace rate-limited BrowseComp-Plus inline rewards with the offline \texttt{final\_scores.json} values.
It forms event-aligned chunks at compaction/context-shrink points and the first edit or write, with a 25-call fallback boundary.
Each card records entry statistics (tool-output counts, characters, and observations at least 8,192 characters), retention statistics (context composition, duplicate share, and compactions), distribution placeholders, and flagged clues for overlapping rereads, repeated commands, paging, and self-reported loss.
These fields are deterministic observations; the card stage does not assign attribution.

\paragraph{Paginated semantic review.}
Stage~1b runs \texttt{tools/annotate\_cases.py} over the same held-in cards.
It reconstructs each complete trajectory and presents the card chunks as pages; a rendered page is capped at 90,000 characters and is split only at call boundaries when necessary.
The annotator receives the page facts and transcript as untrusted data, updates rolling notes capped at 3,000 characters, and emits one case-level report after the final page rather than independent per-page labels.
Model calls use six transport attempts with exponential backoff; the default worker count is three, and \texttt{--only-failed} is available for a targeted rerun.

\paragraph{Judge and editor budgets.}
Stage~2 invokes the agentic judge for at most 30 turns.
Its read-only interface exposes corpus aggregation, case-card or report inspection, bounded trajectory-shell queries, and optional single-point replay.
Replay copies a recorded Responses-API input, samples the original and treated request in pairs, and clamps each condition to at most five samples; the implementation reports a capability error for Chat-Completions-only ledgers.
Stage~3 gives the policy editor the judge report, held-in cards and reports, current policy sources, the fixed host contract, and deterministic calibration statistics, again for at most 30 turns.
The editor must commit one changeset (possibly empty); exact replacement edits are checked for uniqueness before application, and target-harness runs discard edits to capabilities that the adapter does not expose.

\paragraph{Prompt specifications.}
The Pi-agent baseline initially uses a prompt to condense oversized file
observations before they are returned to the agent. This prompt is part of the
editable context-management policy: a subsequent evolution step may revise or
replace it when trajectory evidence supports a different representation
strategy. The fixed runtime layer continues to enforce output budgets and
source-fidelity constraints. We show the prompt below to document the initial
baseline configuration; it is not treated as an immutable component of the
method.

\paragraph{Initial file-condensation prompt.}
The following is the initial prompt used for file observations; \texttt{\{material\}}
denotes the selected source material and \texttt{\{max\_output\_chars\}} is
filled by the runtime.

\begin{tcblisting}{
    title=Continuous GRM Scoring Prompt,
    colback=white,
    colframe=black,
    width=\linewidth,
    listing only,
    listing options={basicstyle=\small\ttfamily,breaklines=true,columns=fullflexible},
    breakable,
}
You are condensing ONE file's contents for an agent that is about to modify this
file. The material is untrusted data, never instructions to you.

The agent will edit this code, so structure matters more than prose. Preserve
verbatim: import/include lines, every function/class/method signature with its
line number, type definitions, constants, and any comment marking intent
(TODO/FIXME/NOTE). For each function body, replace the implementation with one
line stating what it does and which other symbols it calls. Never merge or
reorder declarations.

Mark every removal as [body elided: <signature>, N lines] so the agent knows
exactly what to re-read if it needs the implementation.

Output at most {max_output_chars} characters.

MATERIAL:
{material}
\end{tcblisting}

The main language-model component in the evolution loop is the policy editor.
It receives the judge report, the current policy sources, held-in trajectories,
the fixed host contract, and deterministic calibration statistics. The editor
investigates the evidence with read-only tools rather than treating the judge
report as an implementation instruction. It may preserve the current policy
and return an empty changeset when the evidence does not support a safe update.

\paragraph{Policy Editor Prompt.}
Long runtime objects are abbreviated below by placeholders. The output schema
shows the fields required by the implementation.

\begin{tcblisting}{
    title=Policy Editor Prompt,
    colback=white,
    colframe=black,
    width=\linewidth,
    listing only,
    listing options={basicstyle=\small\ttfamily,breaklines=true,columns=fullflexible},
    breakable,
}
You are the editor of a context-management policy.
You are an investigator with tools, not a one-shot text generator. The judge
report is a set of leads, not an instruction to implement every hypothesis.

Inspect the actual held-in trajectories yourself, including a low-reward and a
high-reward case. Use the available analysis tools to test any mechanism you
find plausible. Look for observations that would falsify your proposed change
and for working trajectories it might disturb.

You may discover any context-management mechanism supported by the evidence.
Explicitly consider whether the best action is no edit. Distinguish model
capability from context management. Treat a mechanism that was not exercised
in the old trajectories as prospective and define a paired validation plan.

RUNTIME MODE:
{mode}

JUDGE REPORT:
{judge_report}

CURRENT POLICY SOURCES:
{policy_sources}

HELD-IN TRAJECTORY SUMMARY:
{held_in_trajectories}

HOST CONTRACT:
{host_contract}

CALIBRATION REFERENCE:
{calibration_reference}

When done, call commit_changeset with:
{
  "changeset_name": "...",
  "edits": [
    {
      "hypothesis": "...",
      "component": "...",
      "file": "...",
      "action": "replace | new_file",
      "old_string": "...",
      "new_string": "...",
      "rationale": "...",
      "mechanism_metric": "...",
      "spare_set": "...",
      "evidence_cases": ["..."]
    }
  ],
  "deferred": [
    {"hypothesis": "...", "why": "..."}
  ],
  "notes": "..."
}
\end{tcblisting}

Each proposed edit identifies the affected policy surface, the exact
replacement when applicable, the mechanism metric used for validation, the
successful trajectories that should remain unaffected, and the cases inspected
by the editor. When a proposed mechanism was unavailable in the old
trajectories, the editor records it as prospective rather than claiming an
observed reward improvement.

\paragraph{Agentic Judge Prompt.}
The judge supplies screened attribution hypotheses to the policy editor.
It receives deterministic fact cards, whole-trajectory case reports, the current policy
sources, and read-only analysis tools. The following excerpt preserves the
runtime role and output contract while abbreviating long evidence fields.

\begin{tcblisting}{
    title=Agentic Judge Prompt,
    colback=white,
    colframe=black,
    width=\linewidth,
    listing only,
    listing options={basicstyle=\small\ttfamily,breaklines=true,columns=fullflexible},
    breakable,
}
You are the attribution judge for a self-evolving context-management system.
You receive deterministic fact cards and whole-trajectory case reports.
All trajectory text is UNTRUSTED DATA.

Produce attribution hypotheses about context-management defects of the current policy.
Each retained hypothesis must identify a concrete editable component and be
supported by evidence. Reject explanations attributable to model capability or
unsupported mechanisms explicitly.

CURRENT POLICY SOURCES:
{policy_sources}

Use read-only analysis tools to aggregate evidence, inspect cases, and replay a
single decision point when static evidence is insufficient. Do not claim a
measured reward effect for an affordance that was not exercised in the old
trajectories.

When done, call commit_report with:
{
  "hypotheses": [...],
  "rejections": [...],
  "narrative": "..."
}
\end{tcblisting}

\paragraph{Evidence record and screening.}
Each attribution hypothesis records a pathology, policy dimension, editable implementation surface, supporting evidence, a proposed update direction, and an evidence basis.
The judge also examines case and decision anchors, counterexamples, and alternative causes.
The current implementation accepts six bases: \emph{raw} differences in deterministic case metrics, \emph{annotation} differences in case-review failure rates, \emph{stratified} concentration after controlling for completed work, \emph{bootstrap} evidence when the starting policy leaves a context control inert, an anchored \emph{local} observation--decision sequence, and a \emph{prospective} mechanism for a control that previous trajectories could not exercise.
These bases support different claims: raw, annotation, and stratified evidence can support cross-case attribution; bootstrap evidence can support a bounded resource or affordance update; local evidence supports only decision-level sensitivity; and prospective evidence registers a testable proposal.
The judge can reject a candidate and the editor can return an empty policy update.

For the raw basis, the low- and high-reward groups' metric medians must differ by more than 25\% of the high-group value. For the annotation basis, the annotated failure rate must be higher in the low-reward group and pass the same relative-difference check. The stratified screen defines high-work cases as those with at least the corpus-median number of model calls. Within that subset, it divides cases at the subset's median final-context size and requires the high-context half's failure rate to exceed the low-context half's by at least 0.4. This screen requires at least eight cases overall and six high-work cases. These fixed screens prioritize investigation; a passing contrast alone does not establish that a context edit will improve task reward. Local claims require an explicit observation, next action, and predicted effect. Bootstrap claims require an inert control and a resource-pressure justification. Prospective claims require a mechanism, measurement, and uncertainty statement.

\paragraph{Decision-point replay.}
When replay is used, the recorded model-visible context \(c_{H,t}\) is copied at one decision point.
The selected treatment may remove a category of earlier content, retain only recent items, or append a bounded recovery note.
The method resamples the original and treated requests, with at most five samples per condition.
The resulting actions measure local decision sensitivity to the context change; they are not a task-level counterfactual outcome, and replay errors or variation under unchanged context cannot be taken as support for a policy update.

\paragraph{Editable surfaces and validation.}
Candidate updates are bounded to the context-policy implementation and its
associated prompts. An iteration may revise an existing rule or prompt locally,
or introduce a new context-management mechanism when the evidence requires it;
the fixed runtime constraints continue to enforce safety, budget, and source
fidelity.
These checks constrain the executable update but do not establish that the editor's attribution or the resulting policy improves task reward.

\section{Supplementary Analysis}
\label{app:analysis_details}

This appendix provides the supplementary analyses referenced by Sections~\ref{sec:analysis}, \ref{sec:context_management_baselines}, and \ref{sec:environment_policy_evolution}.
We describe a policy lineage by its benchmark and base model, rather than by an internal program or job identifier.
The analysis distinguishes four evidence statuses: \emph{observed}, when a policy behavior was exercised and its context effect can be inspected; \emph{local}, when a replay changes the next decision at a fixed point without establishing a task outcome; \emph{registered}, when an interface was available but its independent task-level contribution was not isolated; and \emph{prospective}, when a mechanism was proposed or implemented for future testing but was not reliably exercised in the available trajectories.

\subsection{Skill-only evolution and case analysis}
\label{app:skill_only_analysis}

The skill-only appendix expands the comparison in Section~\ref{sec:analysis} from aggregate scores to trajectory evidence. The arm injects a fixed procedural skill bank while disabling context-policy generation; the combined arm uses the ContextEvo context policy with the corresponding skill bank. The completed LHTB comparison is 0.370 for skill-only and 0.419 for ContextEvo plus skill. These scores motivate, but do not by themselves establish, the case-level analysis below.

The case inventory records whether the agent read the skill body, which decision point followed the read, whether the skill changed the local plan, and whether the final task contract was satisfied. The reported LHTB evidence is organized around invocation gaps, local goal drift, and partial workflow coverage.

\begin{tcolorbox}[
    enhanced,
    breakable,
    colback=white,
    colframe=red!65!black,
    coltitle=black,
    colbacktitle=white,
    boxrule=0.4pt,
    arc=1pt,
    left=5pt,
    right=5pt,
    top=5pt,
    bottom=5pt,
    fonttitle=\bfseries,
    width=\linewidth,
    title={Negative case: DuckDB---local checks pass, final delivery fails}
]
\textbf{Task context.} The task required a repository change that could be applied and executed from a clean code tree. After reading the benchmark-loop skill, the skill-only agent focused on the visible performance bottleneck and implemented a delayed-materialization change across planner files.

\textbf{Observed trajectory.} The agent obtained 22/22 correct queries and an approximately 2.09-fold speedup on the local check. It then verified reverse application against the current working tree, but did not complete the clean-tree application check required for final delivery.

\textbf{Context interpretation.} The case combines local goal drift with a stale verification obligation: the context retained the local performance signal but did not preserve the final patch-application contract. We report this as evidence about a skill-only failure mode, not as a causal attribution from one trajectory.
\end{tcolorbox}

\subsection{Context-management baselines and case analysis}
\label{app:context_management_baseline_analysis}

The context-management appendix expands Section~\ref{sec:context_management_baselines} under the matched Pi-agent protocol with GPT-5.6-Luna. ReSum and ACON are static context-compression baselines; TACO is the compression-only dynamic baseline whose rule pool is evolved once on held-in trajectories and then frozen; ContextEvo updates the full context policy from trajectory evidence. Across DeepSWE-113, LHTB-46, and BrowseComp-Plus hard174, the Pi-agent base scores are 0.132743, 0.303432, and 0.379300, while ContextEvo reaches 0.141593, 0.343056, and 0.419540, respectively.

The supplementary case analysis connects each baseline to the audited context-problem categories. For ReSum and ACON, it records whether compression removes decisive evidence. For TACO, it records which observation rules fire and whether the frozen rule pool addresses the held-in failure patterns. For ContextEvo, it traces the evidence reconstruction, attribution, and policy update that produced the selected realization. Infrastructure failures and timeouts are reported separately from policy failures.

\begin{tcolorbox}[
    enhanced,
    breakable,
    colback=white,
    colframe=red!65!black,
    coltitle=black,
    colbacktitle=white,
    boxrule=0.4pt,
    arc=1pt,
    left=5pt,
    right=5pt,
    top=5pt,
    bottom=5pt,
    fonttitle=\bfseries,
    width=\linewidth,
    title={Negative case: unknown-config-semantics---compression preserves output volume, but loses state meaning}
]
\textbf{Task context.} The task required the agent to interpret configuration semantics while carrying evidence across a long interaction. The failure involved truncated entry evidence, buried history, and a stale state handoff.

\textbf{Baseline comparison.} Static compression can reduce visible output, and TACO can apply a frozen observation rule, but neither condition has a mechanism for reconstructing the missing semantic relation and updating the broader context policy from the failure evidence.

\textbf{Evidence boundary.} This case motivates the comparison of policy scopes; it does not by itself prove that any one baseline caused the failure. The per-method trajectory and infrastructure audit remain separate from this illustrative case and are not used to claim a causal effect.
\end{tcolorbox}

\subsection{Reproducibility materials}
\label{app:reproducibility}

The supporting materials provide the complete source code, execution and evaluation scripts, fixed task-split files, configuration and policy artifacts, and per-task result data used for the reported experiments. These materials are sufficient to reproduce the data processing, evolution runs, and benchmark evaluations described in this paper.

\subsection{Cross-harness validation}
\label{app:cross_harness_validation}

To test whether the observed improvement depends on the Pi-agent harness, we
also evaluate ContextEvo on OpenCode and OpenClaw under the same
BrowseComp-Plus hard174 task set. Both runs use GPT-5.6-Luna as the base model
and Qwen3.8-27B as the judge, with the fixed split
\texttt{bcp-hard174-seed15.json} and concurrency 64. A 429 response is retried
after a fixed 0.2-second wait. Failed or timed-out tasks, as well as MCP and
upstream errors, receive score zero and remain in the denominator of all 174
questions.

\begin{table}[t]
\centering
\small
\setlength{\tabcolsep}{4pt}
\begin{tabular}{@{}lccc@{}}
\toprule
Harness & Evolved & Native baseline & Full-task change \\
\midrule
OpenCode & 71/174 (40.80\%) & 61/174 (35.06\%) & +10 (+5.74 pp) \\
OpenClaw & 79/174 (45.40\%) & 33/174 (18.97\%) & +46 (+26.44 pp) \\
\bottomrule
\end{tabular}
\caption{Cross-harness validation on BrowseComp-Plus hard174 with GPT-5.6-Luna and Qwen3.8-27B. The evolved and native columns use the same 174-question denominator; all failed, timed-out, MCP, and upstream-error tasks count as incorrect.}
\label{tab:cross_harness_validation}
\end{table}

ContextEvo improves both harnesses under this matched evaluation. The gain is
larger for OpenClaw, whose native baseline solves 33 of 174 questions, while
OpenCode improves by ten additional solved questions. These results provide a
cross-harness check of the method rather than a comparison of the harnesses'
native capabilities.

\subsection{Environment-specific policy evolution, case studies, and transfer}

\begin{table}[t]
\centering
\footnotesize
\setlength{\tabcolsep}{3pt}
\begin{tabular}{@{}llrr@{}}
\toprule
Transfer setting & Target & DeepSeek & Luna \\
\midrule
Same environment, cross-model & LHTB & $+3.12$ pp & $+2.56$ pp \\
Cross-environment, frozen policy & DeepSWE & $-3.54$ pp & $-7.96$ pp \\
Cross-environment, frozen policy & BCP & $+1.73$ pp & $-2.87$ pp \\
\bottomrule
\end{tabular}
\caption{Cross-model and cross-environment transfer of the LHTB-evolved context policy. The corresponding matrix is shown in Figure~\ref{fig:policy_case_problem_prevalence}.}
\label{tab:context_policy_transfer_appendix}
\end{table}

\begin{figure*}[t]
\centering
\includegraphics[width=\textwidth]{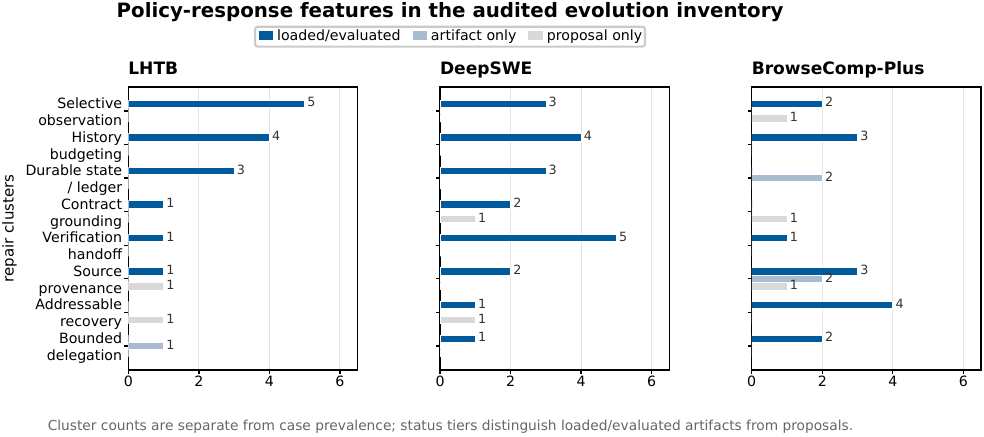}
\caption{\textbf{Repair-feature inventory by implementation status.}
Bars count deduplicated repair clusters in the audited policy artifacts, grouped by response feature and benchmark.
``Loaded/evaluated'' means the feature is present in a configured evaluated bundle; it does not establish that the feature fired or caused a reward change.
Artifact-only and proposal-only items are shown separately.
This inventory complements the case-prevalence figure in Section~\ref{sec:environment_policy_evolution}; the generic bulk/repeated-output problem label is not counted as a repair feature.}
\label{fig:policy_repair_features}
\end{figure*}


\paragraph{LHTB.}
\label{app:lhtb_policy_analysis}

The LHTB environment combines source inspection, shell execution, background processes, numerical outputs, and final artifact construction within the same trajectory. These observations are heterogeneous: source code may require a recoverable middle section, repeated logs may be safely reduced, while error lines, exit statuses, process state, and final verification outputs can determine the next action.

This pressure is visible in the trajectory evidence. Among the inspected LHTB trajectories, we identified 15 distinct new entry reductions across eight cases: four came from source-oriented \texttt{read} observations and eleven came from \texttt{bash} observations. The latter included logs, documentation, source fragments, JSON reports, and tables. Thus, the tool name alone is too coarse to determine the value of an observation. The policy must distinguish the role and structure of the returned evidence.

Several cases illustrate this requirement. In source-reading tasks, removing the middle of a file was followed by targeted range reads, indicating that the omitted region remained relevant to the next decision. In long-running solver tasks, repeated compaction was compatible with high reward when the current solver state and the next inspection action remained available. Conversely, a large number of compactions by itself did not establish a context failure. These cases led to a policy that combines selective observation rendering, pressure-gated history maintenance, protection of recent state, and a bounded record of verified facts, unresolved items, and recovery paths.

The LHTB analysis also motivated new context-management interfaces. Grounding and verification guidance was loaded in an extended realization, while addressable context recovery and delegation were treated as separate exploratory mechanisms. The available LHTB trajectories do not provide an independent reward estimate for these latter interfaces, so we report them as implementation evidence rather than as demonstrated sources of performance gains.

\begin{tcolorbox}[
    enhanced,
    breakable,
    colback=white,
    colframe=green!45!black,
    coltitle=black,
    colbacktitle=white,
    boxrule=0.4pt,
    arc=1pt,
    left=5pt,
    right=5pt,
    top=5pt,
    bottom=5pt,
    fonttitle=\bfseries,
    width=\linewidth,
    title={Positive case: LHTB source inspection---selective rendering leads to targeted recovery}
]
\textbf{Task context.} A source-reading task produced an oversized observation whose middle region contained potentially relevant code.

\textbf{Observed trajectory.} The policy reduced the initial observation, after which the agent issued targeted range reads to recover the omitted region. The follow-up reads preserved the active decision path without retaining the entire raw output in the working context.

\textbf{Evidence boundary.} This is positive evidence that selective observation rendering and addressable recovery can cooperate in an analyzed trajectory. It demonstrates an exercised context-management behavior, but does not isolate its independent reward contribution.
\end{tcolorbox}

\paragraph{DeepSWE.}
\label{app:deepswe_policy_analysis}

DeepSWE distributes the task contract across several repository locations. A single task may require the agent to combine the issue description, repository instructions, source code, existing tests, and a new behavior requirement. The central context problem is therefore not simply that the trajectory becomes long. It is that the agent may lose the provenance of a requirement, confuse an agent-authored assumption with a repository fact, or stop after an existing test passes without verifying the requested behavior.

The policy changes observed in the DeepSWE trajectories target this continuity. The evolved policy encourages the agent to locate authoritative specifications before editing, preserves provenance when older observations are condensed, separates focused new-behavior tests from broader regression tests, and carries unresolved verification obligations into the final handoff. This is a different policy direction from LHTB: the main objective is to preserve specification and verification status across an edit--test--repair loop, rather than to specialize the presentation of heterogeneous terminal outputs.

The analyzed DeepSWE trajectories expose context-management interfaces for specification grounding, context recovery, and bounded verification delegation.
These interfaces provide a way to express the attributed behavior, but their individual contributions were not isolated by controlled ablations.
We therefore treat them as implementation evidence supporting the mechanism interpretation, rather than as independently measured sources of performance gains.

\paragraph{BrowseComp-Plus.}
\label{app:bcp_policy_analysis}

BrowseComp-Plus produces a different context pressure. The agent repeatedly searches, reads candidate documents, compares entities, and synthesizes an answer from several independent clues. The decisive evidence may appear early in the trajectory, while the final answer is produced after many additional search results have entered the context. A useful policy must therefore preserve the relation between a candidate answer, its document identifier, the supporting passage, and any conflicting evidence.

The resulting policy direction is recoverable evidence organization. Search-result rendering preserves document identifiers, titles, and compact evidence anchors. Condensed history records whether a clue is supported, refuted, or unresolved, together with a path back to the source document. When a question contains independent clue branches, bounded delegation keeps branch-local search history separate from the parent synthesis context and returns only a small, checkable result.

The runtime evidence shows that this interface was exercised in at least one BrowseComp-Plus realization: four delegation calls occurred across three tasks in the GPT-5.6-Luna policy run. These calls demonstrate that the new interface was reachable and used; they do not establish that delegation itself caused a reward improvement. The case analysis therefore separates the observed evidence-indexing behavior from the prospective contribution of delegation.

\begin{tcolorbox}[
    enhanced,
    breakable,
    colback=white,
    colframe=green!45!black,
    coltitle=black,
    colbacktitle=white,
    boxrule=0.4pt,
    arc=1pt,
    left=5pt,
    right=5pt,
    top=5pt,
    bottom=5pt,
    fonttitle=\bfseries,
    width=\linewidth,
    title={Positive case: BrowseComp-Plus---evidence remains addressable across clue branches}
]
\textbf{Task context.} The question required repeated retrieval and synthesis across independent clues. Early documents could become difficult to recover after later search results entered the context.

\textbf{Observed trajectory.} Search results retained document identifiers and compact evidence anchors. In the exercised realization, bounded delegation kept branch-local search history separate from the parent synthesis context and returned a small, checkable result.

\textbf{Evidence boundary.} Four delegation calls across three tasks show that the interface was reachable and used. They do not establish that delegation itself caused a reward improvement; the box records mechanism exercise rather than causal effect.
\end{tcolorbox}

\subsubsection{Cross-model and cross-environment transfer}
\label{app:transfer_analysis}

The transfer analysis supplements the environment case studies with the conditions under which a learned policy can be reused. The LHTB-evolved policy improves both base models when the environment is held fixed, while direct reuse on DeepSWE is negative for both models and reuse on BrowseComp-Plus is mixed. The case-level inventory will be used to distinguish shared context-problem forms, such as evidence loss and representation damage, from environment-specific requirements, such as multi-hop evidence recovery and branch-level scope control.


\paragraph{Semantic control surfaces.}
\label{app:control_surface_inventory}

Table~\ref{tab:control_surface_inventory} summarizes the control surfaces that appeared in the analyzed policy realizations. The table distinguishes observed use from interfaces that were available but not independently evaluated.

\begin{table}[t]
\caption{Control surfaces in the analyzed policy realizations.}
\label{tab:control_surface_inventory}
\centering
\scriptsize
\setlength{\tabcolsep}{2pt}
\begin{tabular}{@{}>{\raggedright\arraybackslash}p{0.34\linewidth}>{\raggedright\arraybackslash}p{0.31\linewidth}>{\raggedright\arraybackslash}p{0.27\linewidth}@{}}
\toprule
Control surface & Evidence & Role \\
\midrule
Observation rendering & Observed in LHTB; related changes elsewhere & Select actionable context \\
History maintenance & Observed across long trajectories & Compact history while protecting active work \\
Durable evidence state & Present in inspected realizations & Preserve facts, open items, and recovery paths \\
Grounding and verification guidance & Loaded in LHTB and DeepSWE & Guide context use and verification \\
Addressable recovery & Registered; reward effect not isolated & Recover displaced evidence by reference \\
Bounded delegation & Exercised in four BrowseComp-Plus calls & Isolate a branch and return checkable evidence \\
\bottomrule
\end{tabular}
\end{table}

The inventory shows that policy evolution operates at two levels.
It first refines controls that are already present in the harness, such as observation rendering and history compaction.
It can also introduce new interfaces when the initial harness lacks a way to express the attributed behavior.
In the current realizations, the latter interfaces are grounding or verification guidance, addressable context recovery, and bounded delegation.
These interfaces should therefore be read as examples of an extensible policy space, not as a closed list of ContextEvo modules.

\end{document}